\documentclass[10pt,twocolumn,letterpaper]{article}

\usepackage[pagenumbers]{cvpr} 

\usepackage{multirow}

\usepackage[dvipsnames]{xcolor}

\definecolor{cvprblue}{rgb}{0.21,0.49,0.74}
\usepackage[pagebackref,breaklinks,colorlinks,citecolor=cvprblue]{hyperref}
\PassOptionsToPackage{hyphens}{url}\usepackage{hyperref}

\newcommand{\smallurl}[1]{{\footnotesize\url{#1}}}

\def\method{TripoSR: Fast 3D Object Reconstruction from a Single Image\xspace}
\def\methodshort{TripoSR\xspace}

\def\paperID{*****} 
\def\confName{CVPR}
\def\confYear{2024}

\title{\method}

\author{
Dmitry Tochilkin$^{1}$ \quad David Pankratz$^{1}$ \quad Zexiang Liu$^{2}$ \quad Zixuan Huang$^{1}$ \quad Adam Letts$^{1}$ \quad
\\ Yangguang Li$^{2}$ \quad Ding Liang$^{2}$ \quad Christian Laforte$^{1}$ \quad Varun Jampani$^{1*}$ \quad Yan-Pei Cao$^{2*}$
\\ \\
$^1$Stability AI, $^2$Tripo AI
}

\begin{document}
\gdef\UrlBreaks{\do\/\do\-\do\s\do\b\do\l\do\r\do\c\do\h\do\t\do\y\do\a\do\i}
\sloppy

\twocolumn[{%
\renewcommand\twocolumn[1][]{#1}%
\maketitle
\begin{center}
    \centering
    \captionsetup{type=figure}
    \includegraphics[width=1\textwidth]{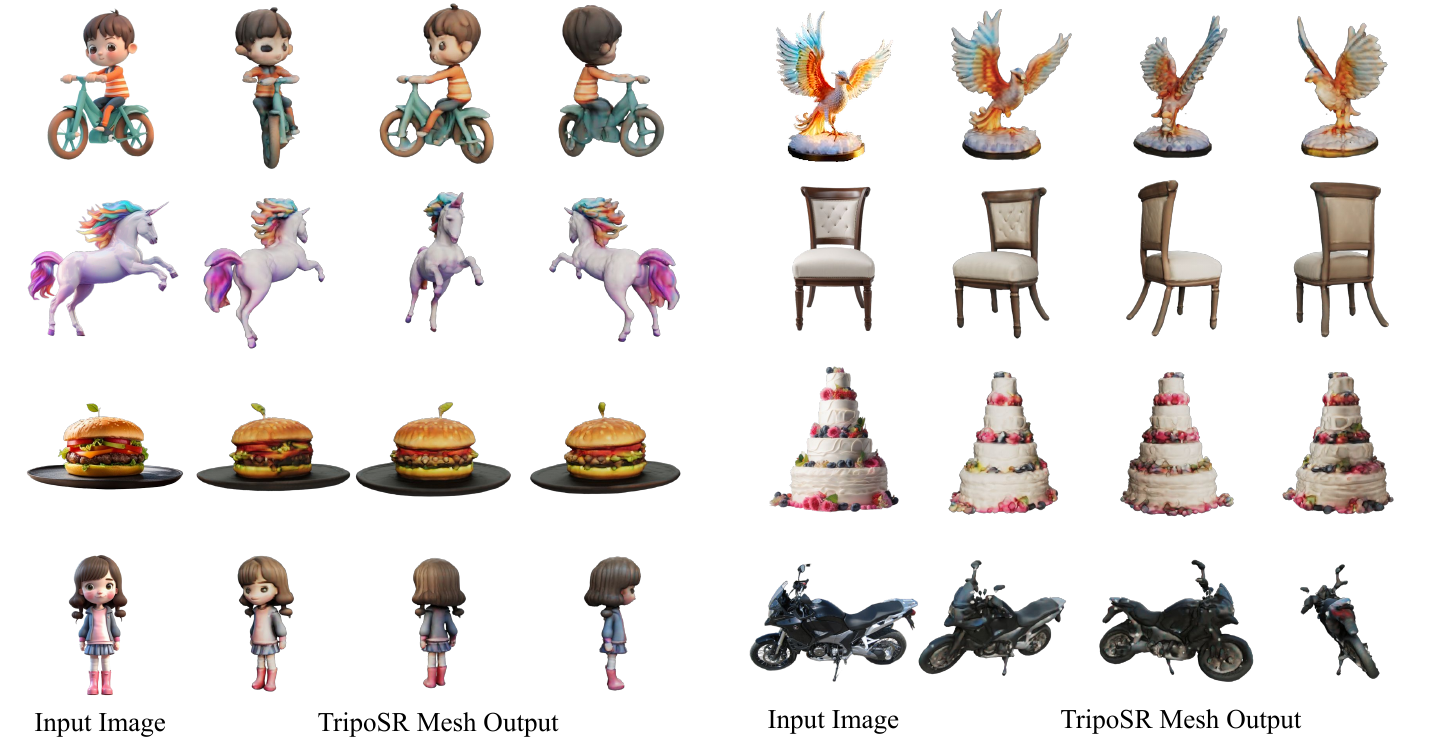}
    \captionof{figure}{We present TripoSR, a 3D reconstruction model that reconstructs high-quality 3D from single images in under 0.5 seconds. Our model achieves state-of-the-art performance and generalizes to objects of various types and input images across different domains.}
\label{fig:teaser-results}
\end{center}%
}]

\def\thefootnote{*}\footnotetext{Equal advising.}\def\thefootnote{\arabic{footnote}}

\begin{abstract}
This technical report introduces \methodshort, a 3D reconstruction model leveraging transformer architecture for fast feed-forward 3D generation, producing 3D mesh from a single image in under 0.5 seconds. Building upon the LRM~\cite{hong2023lrm} network architecture, \methodshort integrates substantial improvements in data processing, model design, and training techniques. Evaluations on public datasets show that \methodshort exhibits superior performance, both quantitatively and qualitatively, compared to other open-source alternatives. Released under the MIT license, \methodshort is intended to empower researchers, developers, and creatives with the latest advancements in 3D generative AI.

\small{
\noindent\textbf{Model:} \url{https://huggingface.co/stabilityai/TripoSR} \\
\noindent\textbf{Code:} \url{https://github.com/VAST-AI-Research/TripoSR} \\
\noindent\textbf{Demo:} \url{https://huggingface.co/spaces/stabilityai/TripoSR}
}

\end{abstract}    
\vspace{-5mm}
\section{Introduction}
\label{sec:intro}
The landscape of 3D Generative AI has witnessed a confluence of developments in recent years, blurring the lines between 3D reconstruction from single or few views and 3D generation~\cite{zhou2023sparsefusion,zou2023sparse3d,wu2023reconfusion,chan2023genvs,hong2023lrm,zou2023triplane,huang2023zeroshape,liu2023zero,guo2023threestudio}. This convergence has been significantly accelerated by the introduction of large-scale public 3D datasets~\cite{deitke2023objaverse,deitke2024objaverse} and advances in generative model architectures. Comprehensive reviews of these technologies can be found in the literature such as ~\cite{li2024advances} and ~\cite{shi2022deep}.

To overcome the scarcity of 3D training data, recent efforts have explored utilizing 2D diffusion models to create 3D assets from text prompts~\cite{poole2022dreamfusion,wang2024prolificdreamer,shi2023mvdream} or input images~\cite{liu2023zero,tang2023dreamgaussian}. DreamFusion~\cite{poole2022dreamfusion}, a notable example, introduced score distillation sampling (SDS), employing a 2D diffusion model to guide the optimization of 3D models. This approach represents a pivotal strategy in leveraging 2D priors for 3D generation, achieving breakthroughs in generating detailed 3D objects. However, these methods typically face limitations with slow generation speed, due to the extensive optimization and computational demands, and the challenge of precisely controlling the output models.

On the contrary, feed-forward 3D reconstruction models achieve significantly higher computational efficiency~\cite{groueix2018papier,gkioxari2019mesh,zhang2018learning,wang2018pixel2mesh,mescheder2019occupancy,Huang_2023_CVPR,hong2023lrm,li2023instant3d,xu2023dmv3d,wang2023pf,liu2023zero,zou2023triplane,huang2023zeroshape,wu2023multiview,tang2024lgm}. Several recent approaches~\cite{hong2023lrm,li2023instant3d,xu2023dmv3d,wang2023pf,liu2023zero,zou2023triplane,huang2023zeroshape,wu2023multiview,tang2024lgm} along this direction have shown promise in scalable training on diverse 3D datasets. These approaches facilitate rapid 3D model generation through fast feed-forward inference and are potentially more capable of providing precise control over the generated outputs, marking a notable shift in the efficiency and applicability of these models.

In this work, we introduce \methodshort model for fast feed-forward 3D generation from a single image that takes less than 0.5 seconds on an A100 GPU. Building upon the LRM~\cite{hong2023lrm} architecture, we introduce several improvements in terms of data curation and rendering, model design and training techniques. Experimental results demonstrate superior performance, both quantitatively and qualitatively, compared to other open-source alternatives. Figure~\ref{fig:teaser-results} shows some sample results of the \methodshort. \methodshort is made available under the MIT license, accompanied by source code, the pretrained model, and an interactive online demo. The release aims to enable researchers, developers, and creatives to advance their work with the latest advancements in 3D generative AI, promoting progress within the wider domains of AI, computer vision, and computer graphics. Next, we introduce the technical advances in our \methodshort model, followed by the quantitative and qualitative results on two public datasets.

\section{\methodshort: Data and Model Improvements}
\label{sec:model}

The design of \methodshort is based on the LRM~\cite{hong2023lrm}, with a series of technical advancements in data curation, model and training strategy. We now give an overview of the model followed by our technical improvements.

\subsection{Model Overview}
Similar to LRM~\cite{hong2023lrm},~\methodshort leverages the transformer architecture and is specifically designed for single-image 3D reconstruction. It takes a single RGB image as input and outputs a 3D representation of the object in the image. 
The core of~\methodshort includes components: an image encoder, an image-to-triplane decoder, and a triplane-based neural radiance field (NeRF).

\begin{table*}
\small
\begin{center}
\begin{tabular}{l l r}
\toprule
\multicolumn{2}{l}{\textbf{Parameter}} & \textbf{Value} \\ 
\midrule
\multirow{4}{*}{\textbf{Image Tokenizer}} & \texttt{image resolution} & $512\times512$ \\
& \texttt{patch size} & $16$ \\
& \texttt{\# attention layers} & $12$ \\
& \texttt{\# feature channels} & $768$ \\
\midrule
\multirow{2}{*}{\textbf{Triplane Tokenizer}} & \texttt{\# tokens} & $32\times32\times3$ \\
& \texttt{\# channels} & $16$ \\
\midrule
\multirow{5}{*}{\textbf{Backbone}} & \texttt{\# channels} & $1024$ \\
& \texttt{attention layers} & $16$ \\
& \texttt{\# attention heads} & $16$ \\
& \texttt{attention head dim} & $64$ \\
& \texttt{cross attention dim} & $768$ \\
\midrule
\multirow{4}{*}{\textbf{Triplane Upsampler}} & \texttt{factor} & $2$ \\
& \texttt{\# input channels} & $1024$ \\
& \texttt{\# output channels} & $40$ \\
& \texttt{output shape} & $64\times64\times40$ \\
\midrule
\multirow{3}{*}{\textbf{NeRF MLP}} 
& \texttt{width} & $64$ \\
& \texttt{\# layers} & $10$ \\
& \texttt{activation} & $\mathrm{SiLU}$ \\
\midrule
\multirow{4}{*}{\textbf{Renderer}} & \texttt{\# samples per ray} & $128$ \\
& \texttt{radius} & $0.87$ \\
& \texttt{density activation} & $\mathrm{exp}$ \\
& \texttt{density bias} & $-1.0$ \\
\midrule
\midrule
\multirow{6}{*}{\textbf{Training}} & \texttt{learning rate} & $4\mathrm{e}{-4}$ \\
& \texttt{optimizer} & AdamW \\
& \texttt{lr scheduler} & CosineAnnealingLR \\
& \texttt{\# warm-up steps} & $2,000$ \\ 
& $\lambda_\mathrm{LPIPS}$ & $2.0$ \\ 
& $\lambda_\mathrm{mask}$ & $0.05$ \\ 
\bottomrule
\end{tabular}
\caption{Model configuration of \methodshort.}
\label{tab:architecture}
\end{center}
\end{table*}

The image encoder is initialized with a pre-trained vision transformer model, DINOv1~\cite{caron2021emerging}, which projects an RGB image into a set of latent vectors. These vectors encode the global and local features of the image and include the necessary information to reconstruct the 3D object.

The subsequent image-to-triplane decoder transforms the latent vectors onto the triplane-NeRF representation~\cite{chan2022efficient}. The triplane-NeRF representation is a compact and expressive 3D representation, well-suited for representing objects with complex shapes and textures. Our decoder consists of a stack of transformer layers, each with a self-attention layer and a cross-attention layer. The self-attention layer allows the decoder to attend to different parts of the triplane representation and learn relationships between them. The cross-attention layer allows the decoder to attend to the latent vectors from the image encoder and incorporate global and local image features into the triplane representation. Finally, the NeRF model consists of a stack of multilayer perceptrons (MLPs), which are responsible for predicting the color and density of a 3D point in space. 

Instead of conditioning the image-to-triplane projection on camera parameters, we have opted to allow the model to ``guess'' the camera parameters (both extrinsics and intrinsics) during training and inference. This is to enhance the model's robustness to in-the-wild input images at inference time. By foregoing explicit camera parameter conditioning, our approach aims to cultivate a more adaptable and resilient model capable of handling a wide range of real-world scenarios without the need for precise camera information.

The architecture's main parameters, such as the number of layers in the transformer, the dimensions of the triplanes, the specifics of the NeRF model, and the main training configurations, are detailed in Table~\ref{tab:architecture}. 
Compared to LRM~\cite{hong2023lrm}, \methodshort introduces several technical improvements which we discuss next. 

\subsection{Data Improvements}
Recognizing the critical importance of data, we have incorporated two improvements in our training data collection: 
\begin{itemize}
    \item \textbf{Data Curation}: By selecting a carefully curated subset of the Objaverse~\cite{deitke2023objaverse} dataset, which is available under the CC-BY license, we have enhanced the quality of training data.
    \item \textbf{Data Rendering}: We have adopted a diverse array of data rendering techniques that more closely emulate the distribution of real-world images, thereby enhancing the model's ability to generalize, even when trained exclusively with the Objaverse dataset.
\end{itemize}

\subsection{Model and Training Improvements}
Our adjustments aim to boost both the model's efficiency and its performance.

\noindent \textbf{Triplane Channel Optimization.}
The configuration of channels within the triplane-NeRF representation plays an important role in managing the GPU memory footprint during both training and inference, due to the high computational cost of volume rendering. Moreover, the channel count significantly influences the model's capacity for detailed and high-fidelity reconstruction. In pursuit of an optimal balance between reconstruction quality and computational efficiency, experimental evaluations led us to adopt a configuration of $40$ channels. This choice enables the use of larger batch sizes and higher resolutions during the training phase, while concurrently minimizing the memory requirements during inference. 

\noindent \textbf{Mask Loss.}
We incorporated a mask loss function during training that significantly reduces ``floater'' artifacts and improves the fidelity of reconstructions:
\begin{align}
\mathcal{L}_\mathrm{mask}(\boldsymbol{\hat{M}}_{v}, \boldsymbol{M}^{\mathit{GT}}_{v}) &= \mathrm{BCE}(\boldsymbol{\hat{M}}_{v}, \boldsymbol{M}^{\mathit{GT}}_{v}),
\label{eq:loss_mask}
\end{align}
where $\boldsymbol{\hat{M}}_{v}$ and $\boldsymbol{M}^{\mathit{GT}}_{v}$ are rendered and ground-truth mask images of the $v$-th supervision view, respectively. The full training loss we minimized during training is: 
\begin{equation}
\begin{split}
\mathcal{L}_\mathrm{recon}(\boldsymbol{I}) &= \frac{1}{V} {\sum_{v=1}^{V}} 
\left( 
\mathcal{L}_\mathrm{MSE}(\boldsymbol{\hat{I}}_{v},\boldsymbol{I}^\mathit{GT}_{v}) \right. \\ 
&+ {\lambda_\mathrm{LPIPS}}\mathcal{L}_\mathrm{LPIPS}(\boldsymbol{\hat{I}}_{v},\boldsymbol{I}^\mathit{GT}_{v}) \\ 
&\left.+ {\lambda_\mathrm{mask}}\mathcal{L}_\mathrm{mask}(\boldsymbol{\hat{M}}_{v}, \boldsymbol{M}^{\mathit{GT}}_{v}) 
\right) 
\end{split}
\label{eq:loss_recon}
\end{equation}

\noindent \textbf{Local Rendering Supervision.} 
Our model fully relies on rendering losses for supervision, thereby imposing a need for high-resolution rendering for our model to learn detailed shape and texture reconstructions.
However, rendering and supervising at high resolutions (e.g., $512\times512$ or higher) can overwhelm computational and GPU memory loads. 
To circumvent this issue, we render $128\times128$-sized random patches from the original $512\times512$ resolution images during training. Crucially, we increase the likelihood of selecting crops that cover foreground regions, thereby placing greater emphasis on the areas of interest. This importance sampling strategy ensures faithful reconstructions of object surface details, effectively balancing computational efficiency and reconstruction granularity.

\begin{figure}[t]
\centering
    \includegraphics[width=1\linewidth]{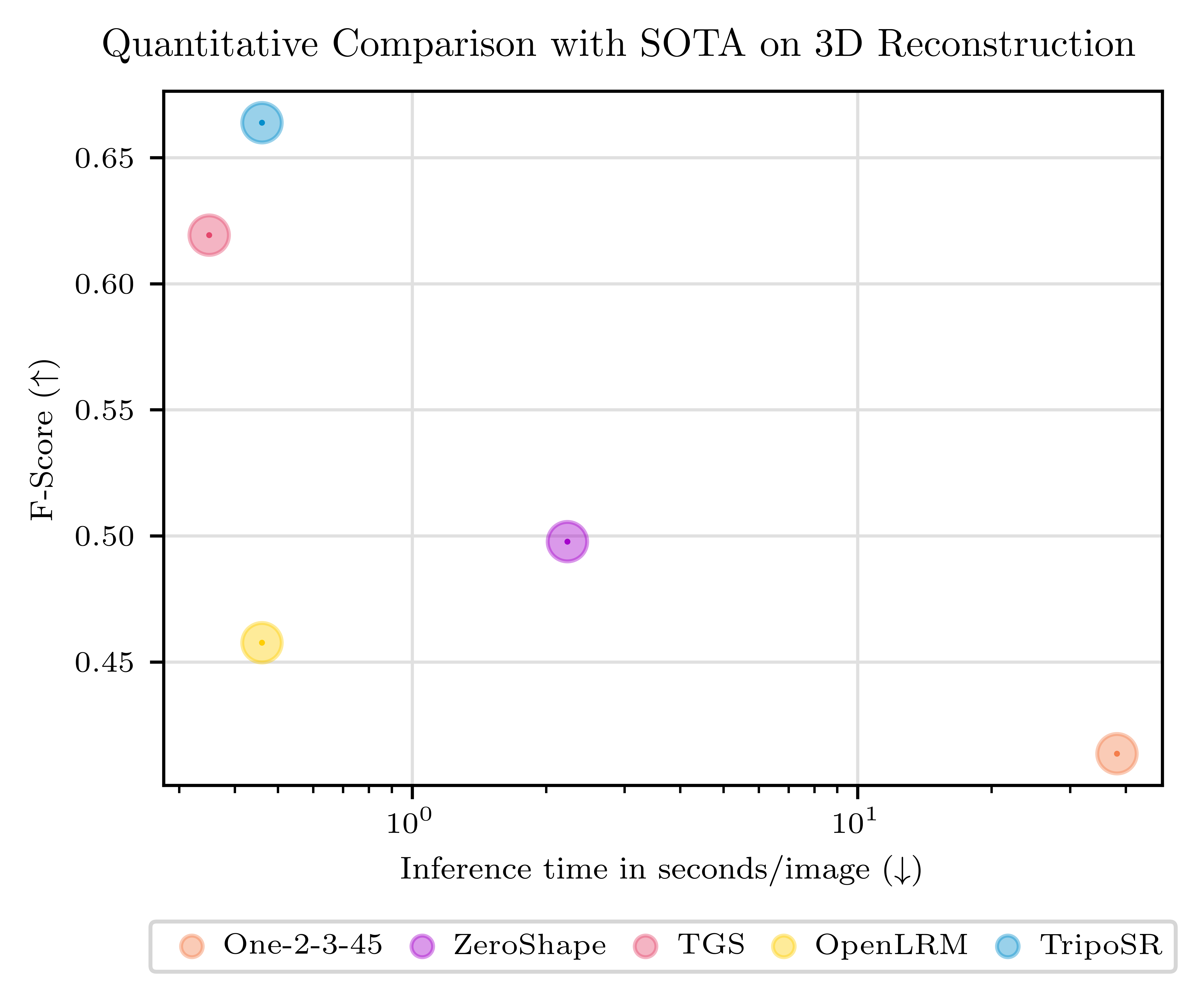}
    \caption{We outperform SOTA methods for 3D reconstruction while achieving fast inference time. In the figure, F-Score with threshold 0.1 is averaged over GSO~\cite{downs2022google} and OmniObject3D~\cite{wu2023omniobject3d}.} 
    \label{fig:scatter-teaser}
\end{figure}

\begin{figure*}[t]
\centering
	\includegraphics[width=\linewidth]{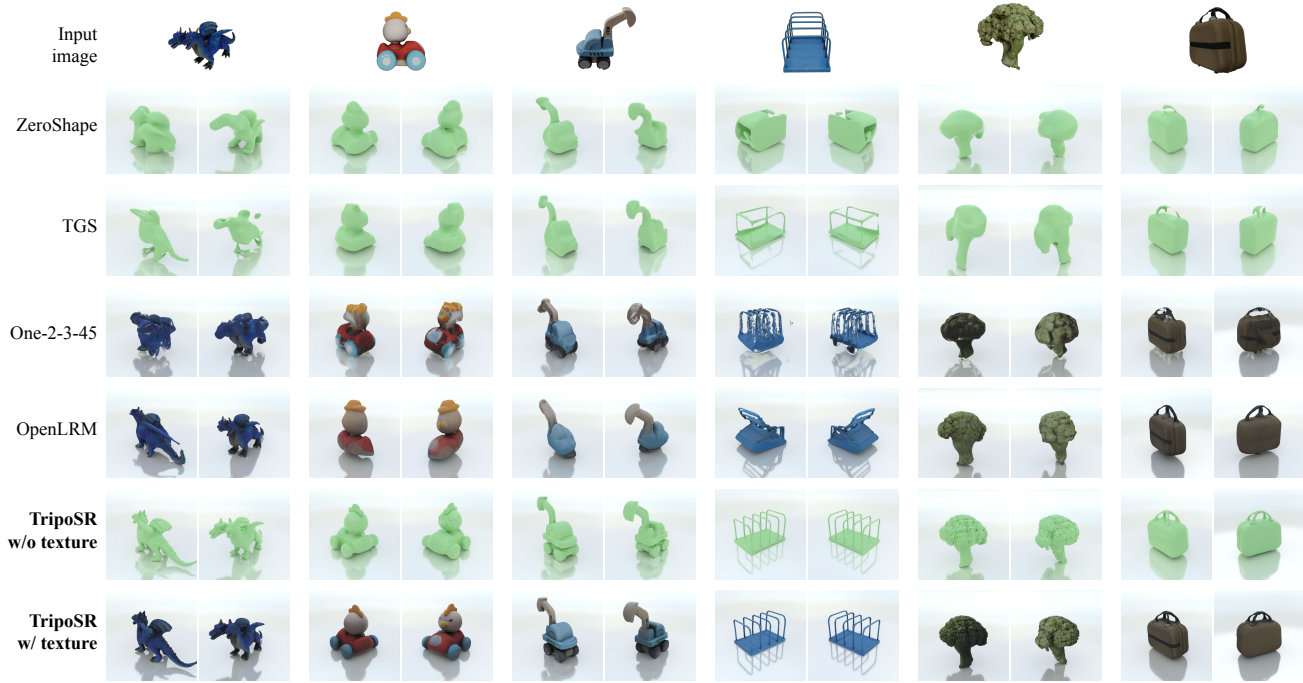}
	\caption{\textbf{Qualitative results.} We compare \methodshort output meshes to other SOTA methods on GSO and OmniObject3D (first four columns are from GSO~\cite{downs2022google}, last two are from OmniObject3D~\cite{wu2023omniobject3d}). Our reconstructed 3D shapes and textures achieve significantly higher quality and better details than previous state-of-the-art methods.}
	\label{fig:qualitative}
 \vspace{-10pt}
\end{figure*}

\section{Results}
\label{sec:results}
We quantitatively and qualitatively compare \methodshort to previous state-of-the-art methods using two different datasets with 3D reconstruction metrics.

\noindent \textbf{Evaluation Datasets}.
We curate two public datasets, GSO~\cite{downs2022google} and OmniObject3D~\cite{wu2023omniobject3d}, for evaluations.
We identify that both datasets include many simple-shaped objects (e.g., box, sphere or cylinder) and can thus cause high validation bias towards these simple shapes. Therefore we manually filter the datasets and select around 300 objects from each dataset to make sure they form a diverse and representative collection of common objects.

\noindent \textbf{3D Shape Metrics}.
We extract the isosurface using Marching Cubes~\cite{10.1145/37402.37422} to convert implicit 3D representations (such as NeRF) into meshes. We sample 10K points from these surfaces to calculate the Chamfer Distance (CD) and F-score (FS). Considering that some methods are not capable of predicting view-centric shapes, we use a brute-force search approach to align the predictions with the ground truth shapes. We linearly search the rotation angle by optimizing for the lowest CD and further employ the Iterative Closest Point (ICP) method to refine the alignment.

\begin{table}
\begin{center}
\scriptsize
\resizebox{\columnwidth}{!}{\begin{tabular}{ l c c c c}
\hline
Method & CD$\downarrow$ & FS@0.1$\uparrow$ & FS@0.2$\uparrow$ & FS@0.5$\uparrow$ \\
\hline
One-2-3-45~\cite{liu2024one}                & 0.227 & 0.382 & 0.630 & 0.878 \\
ZeroShape~\cite{huang2023zeroshape}         & 0.160 & 0.489 & 0.757 & 0.952 \\
TGS~\cite{zou2023triplane}     & 0.122    & 0.637    & 0.846    & 0.968    \\
OpenLRM~\cite{openlrm}                      & 0.180 & 0.430    & 0.698 & 0.938    \\
\methodshort (ours)                              & \textbf{0.111} & \textbf{0.651} & \textbf{0.871} & \textbf{0.980} \\
\hline
\end{tabular}}
\caption{Quantitative comparison of different techniques on GSO~\cite{downs2022google} validation set, where CD and FS refer to Chamfer Distance and F-score respectively.}
\label{tab:volumetric_gso}
\end{center}
\end{table}

\begin{table}
\begin{center}
\scriptsize
\resizebox{\columnwidth}{!}{\begin{tabular}{ l c c c c}
\hline
Method & CD$\downarrow$ & FS@0.1$\uparrow$ & FS@0.2$\uparrow$ & FS@0.5$\uparrow$ \\
\hline
One-2-3-45~\cite{liu2024one}                & 0.197 & 0.445 & 0.698 & 0.907\\
ZeroShape~\cite{huang2023zeroshape}         & 0.144 & 0.507 & 0.786 & 0.968 \\
TGS~\cite{zou2023triplane}     & 0.142    & 0.602    & 0.818    & 0.949    \\
OpenLRM~\cite{openlrm}                      & 0.155    & 0.486    & 0.759    & 0.959    \\
\methodshort (ours)                              & \textbf{0.102} & \textbf{0.677} & \textbf{0.890} & \textbf{0.986} \\
\hline
\end{tabular}}
\caption{Quantitative comparison of different techniques on OmniObject3D~\cite{wu2023omniobject3d} validation set, where CD and FS refers to Chamfer Distance and F-score respectively.}
\label{tab:volumetric_omniobj}
\end{center}
\end{table}

\noindent \textbf{Quantitative Comparisons}.
We compare \methodshort with the existing state-of-the-art baselines on 3D reconstruction that use feed-forward techniques, including One-2-3-45~\cite{liu2024one}, TriplaneGaussian (TGS)~\cite{zou2023triplane}, ZeroShape~\cite{huang2023zeroshape} and OpenLRM~\cite{openlrm}\footnote{We use the openlrm-large-obj-1.0 model.}. As shown in Table~\ref{tab:volumetric_gso} and Table~\ref{tab:volumetric_omniobj}, our TripoSR significantly outperforms all the baselines, both in terms of CD and FS metrics, achieving the new state-of-the-art performance on this task. 

\noindent \textbf{Performance vs. Runtime}.
Another key advantage of~\methodshort is its inference speed. It takes around 0.5 seconds to produce a 3D mesh from a single image on an NVIDIA A100 GPU. Figure~\ref{fig:scatter-teaser} shows a 2D plot of different techniques with inference times along the x-axis and the averaged F-Score along the y-axis. The plot shows that~\methodshort is among the fastest networks, while also being the best-performing feed-forward 3D reconstruction model.

\noindent \textbf{Visual Results}.
We further show the qualitative results of different approaches in Figure~\ref{fig:qualitative}. Because some methods do not reconstruct textured meshes, we render TripoSR reconstructions both with and without vertex color for a better comparison. As shown in the figure, ZeroShape tends to predict over-smoothed shapes. TGS reconstructs more surface details but these details sometimes do not align with the input. Moreover, both ZeroShape and TGS cannot output textured meshes directly~\footnote{TGS leverages 3DGS to represent 3D objects. We follow the paper and utilize their auxiliary point cloud outputs to reconstruct the surface. However, it is non-trivial to reconstruct textures on meshes, (e.g., directly taking vertex colors from the nearest Gaussian leads to noisy textures).}. On the other hand, One-2-3-45 and OpenLRM predict textured meshes, but their estimated shapes are often inaccurate. Compared to these baselines, TripoSR demonstrates a high reconstruction quality for both shape and texture. Our model not only captures a better overall 3D structure of the object, but also excels at modeling several intricate details.

\section{Conclusion}
\label{sec:conclusion}

In this report, we present an open-source feedforward 3D reconstruction model, \methodshort. The core of our model is a transformer-based architecture developed upon the LRM network~\cite{hong2023lrm}, together with substantial technical improvements along multiple axes. Evaluated on two public benchmarks, our model demonstrates state-of-the-art reconstruction performance with high computational efficiency. We hope \methodshort empowers researchers and developers in developing more advanced 3D generative AI models.

\section*{Acknowledgements}
\label{sec:ack}
\noindent \textbf{Tripo AI}. We extend our sincere gratitude to Yuan-Chen Guo and Zi-Xin Zou for their critical roles in coding, demo development, and experimentation.
We also thank Peng Wang for his insightful discussions, Dehu Wang for preparing the datasets for model validation, and Sienna Huang for managing communication within our collaboration.

\vspace{3mm}
\noindent \textbf{Stability AI}. We thank Emad Mostaque, Anel Islamovic, Bryce Wilson, Ana Guillen, Adam Chen, Christian Dowell and Ella Irwin for their help in various aspects of the model development, collaboration and release. We also thank Vikram Voleti for helpful discussions and Eric Courtemanche for his help with visual results. 
{
    \small
    \bibliographystyle{ieeenat_fullname}
    \bibliography{sections/reference}
}


\end{document}